\documentclass{sig-alternate-05-2015}
\newcommand{\tool}{subgraph2vec}
\usepackage{amsmath,amssymb,amsfonts}
\usepackage{multirow}
\usepackage[table,xcdraw]{xcolor}
\usepackage[utf8]{inputenc}
\usepackage{cleveref}
\crefname{section}{§}{§§}
\Crefname{section}{§}{§§}
\usepackage{float}
\usepackage{graphicx}
\usepackage{hyperref}
\usepackage{color}
\hypersetup{
	colorlinks=true,
	linkcolor=red,
	citecolor=blue,
	filecolor=black,
	urlcolor=blue,
}
\usepackage{enumitem}
\makeatletter
\DeclareUrlCommand\ULurl@@{%
	\def{\scriptsize ($ \pm 0.00 $)}Font{\ttfamily\color{blue}}%
	\def{\scriptsize ($ \pm 0.00 $)}Left{\uline\bgroup}%
	\def{\scriptsize ($ \pm 0.00 $)}Right{\egroup}}
\def\ULurl@#1{\hyper@linkurl{\ULurl@@{#1}}{#1}}
\DeclareRobustCommand*\ULurl{\hyper@normalise\ULurl@}
\makeatother

\usepackage{array}
\usepackage{pifont}
\newcolumntype{L}[1]{>{\raggedright\let\newline\\\arraybackslash\hspace{0pt}}m{#1}}
\newcolumntype{C}[1]{>{\centering\let\newline\\\arraybackslash\hspace{0pt}}m{#1}}
\newcolumntype{R}[1]{>{\raggedleft\let\newline\\\arraybackslash\hspace{0pt}}m{#1}}

\usepackage[font=small,labelfont=bf,
justification=justified,
format=plain]{caption}
\DeclareCaptionType{copyrightbox}
\usepackage{amsfonts}
\usepackage{fixltx2e}
\usepackage{bm}
\usepackage{amsmath}
\usepackage{graphicx}
\usepackage{amssymb}
\usepackage{tikz}
\usepackage{array}
\usepackage{xcolor}
\usepackage[mathscr]{euscript}
\usepackage{mathrsfs}
\usepackage{tikz}
\newcommand{\pie}[1]{%
	\begin{tikzpicture}
	\draw (0,0) circle (1ex);\fill (1ex,0) arc (0:#1:1ex) -- (0,0) -- cycle;
	\end{tikzpicture}%
}

\usetikzlibrary{shapes.geometric, arrows}
\tikzstyle{ADG} = [ellipse, minimum width=1cm, minimum height=.75cm,text centered, draw=black, fill=pink!30]
\tikzstyle{PDG} = [ellipse, minimum width=1cm, minimum height=.75cm,text centered, draw=black, fill=yellow!20]
\tikzstyle{SSCFP} = [ellipse, minimum width=1cm, minimum height=.75cm,text centered, draw=black, fill=green!10]
\tikzstyle{Ins} = [ellipse, minimum width=1cm, minimum height=.75cm,text centered, draw=black, fill=brown!10]
\tikzstyle{Signs} = [ellipse, minimum width=1cm, minimum height=.75cm,text centered, draw=black, fill=orange!10]

\usepackage{listings}
\lstdefinestyle{customc}{
	belowcaptionskip=1\baselineskip,
	breaklines=true,
	frame=L,
	xleftmargin=\parindent,
	language=Java,
	showstringspaces=false,
	basicstyle=\footnotesize\ttfamily,
	keywordstyle=\bfseries\color{green!40!black},
	commentstyle=\itshape\color{purple!40!black},
	identifierstyle=\color{blue},
	stringstyle=\color{orange},
}

\usepackage{pifont}
\usepackage[ruled,vlined,linesnumbered]{algorithm2e}

\usepackage{verbatim}

\usepackage{enumerate}
\usepackage{enumitem}
\usepackage[space]{cite}
\usepackage{footmisc}
\usepackage{footnote}
\makeatletter
\newtoks\therules
\therules={}
\def\appendto#1#2{\expandafter#1\expandafter{\the#1#2}}
\def\gobblefirst#1{
	#1\expandafter\expandafter\expandafter{\expandafter\@gobble\the#1}}%
\def\LState{\State\unskip\the\therules}
\def\printindent{\unskip\the\therules}%

\begin{document}

\title{\tool: Learning Distributed Representations 	of Rooted Sub-graphs from Large Graphs}
\numberofauthors{1}
\author{
	\alignauthor
	Annamalai Narayanan$^\dag$, Mahinthan Chandramohan$^\dag$, Lihui Chen$^\dag$, Yang Liu$^\dag$ and Santhoshkumar Saminathan$^\S$\\
	\affaddr{$^\dag$Nanyang Technological University, Singapore}\\
	\affaddr{$^\S$BigCommerce, California, USA}\\
	\email{\texttt{annamala002@e.ntu.edu.sg, \{mahinthan,elhchen,yangliu\}@ntu.edu.sg, santhosh.kumar@yahoo.com}}\\
}
\maketitle
 
\begin{abstract}
In this paper, we present \tool, a novel approach for learning latent representations of rooted subgraphs from large graphs inspired by recent advancements in Deep Learning and Graph Kernels. These latent representations encode semantic substructure dependencies in a continuous vector space, which is easily exploited by statistical models for tasks such as graph classification, clustering, link prediction and community detection. \tool{} leverages on local information obtained from neighbourhoods of nodes to \textit{learn} their latent representations in an \textit{unsupervised} fashion. 
We demonstrate that subgraph vectors learnt by our approach could be used in conjunction with classifiers such as CNNs, SVMs and relational data clustering algorithms to achieve significantly superior accuracies. Also, we show that the subgraph vectors could be used for building a deep learning variant of Weisfeiler-Lehman graph kernel. 
Our experiments on several benchmark and large-scale real-world datasets reveal that \tool{} achieves significant improvements in accuracies over existing graph kernels on both supervised and unsupervised learning tasks. Specifically, on two real-world program analysis tasks, namely, code clone and malware detection, \tool{} outperforms state-of-the-art kernels by more than 17\% and 4\%, respectively.
\end{abstract}

%
%
%

%
%

%
%
\printccsdesc


\keywords{Graph Kernels, Deep Learning, Representation Learning

\section {Introduction}
\label{sec:intro}
Graphs offer a rich, generic and natural way for representing structured data.
In domains such as computational biology, chemoinformatics, social network analysis and program analysis, we are often interested in computing similarities between graphs to cater domain-specific applications such as protein function prediction,  drug toxicity prediction and malware detection.

\textbf{Graph Kernels.} Graph Kernels are one of the popular and widely adopted approaches to measure similarities among graphs \cite{GKVishy,WLK,dgk,sp,gksmoothing}. A Graph kernel measures the similarity between a pair of graphs by recursively decomposing them into \textit{atomic} substructures (e.g., walk \cite{GKVishy}, shortest paths \cite{sp}, graphlets \cite{dgk} etc.) and defining a similarity function over the substructures (e.g., number of common substructures across both graphs). This makes the kernel function correspond to an inner product over substructures in reproducing kernel Hilbert space (RKHS).  
Formally, for a given graph $G$, let $\Phi(G)$ denote a vector which contains counts of atomic substructures, and $\langle \cdot,\cdot \rangle_H$ denote a dot product in a RKHS \textit{H}. Then, the kernel between two graphs $ G $ and $ G' $ is given by 
\begin{equation}
K (G, G') = \langle \Phi (G), \Phi (G') \rangle_H
\end{equation}

From an application standpoint, the kernel matrix $\mathfrak{K}$ that represents the pairwise similarity of graphs in the dataset (calculated using eq. (1)) could be used in conjunction with kernel classifiers (e.g., Support Vector Machine (SVM)) and relational data clustering algorithms to perform graph classification and clustering tasks, respectively.

\begin{figure}[t]
	\centering
	\includegraphics[height=4.2cm,width=7.5cm]{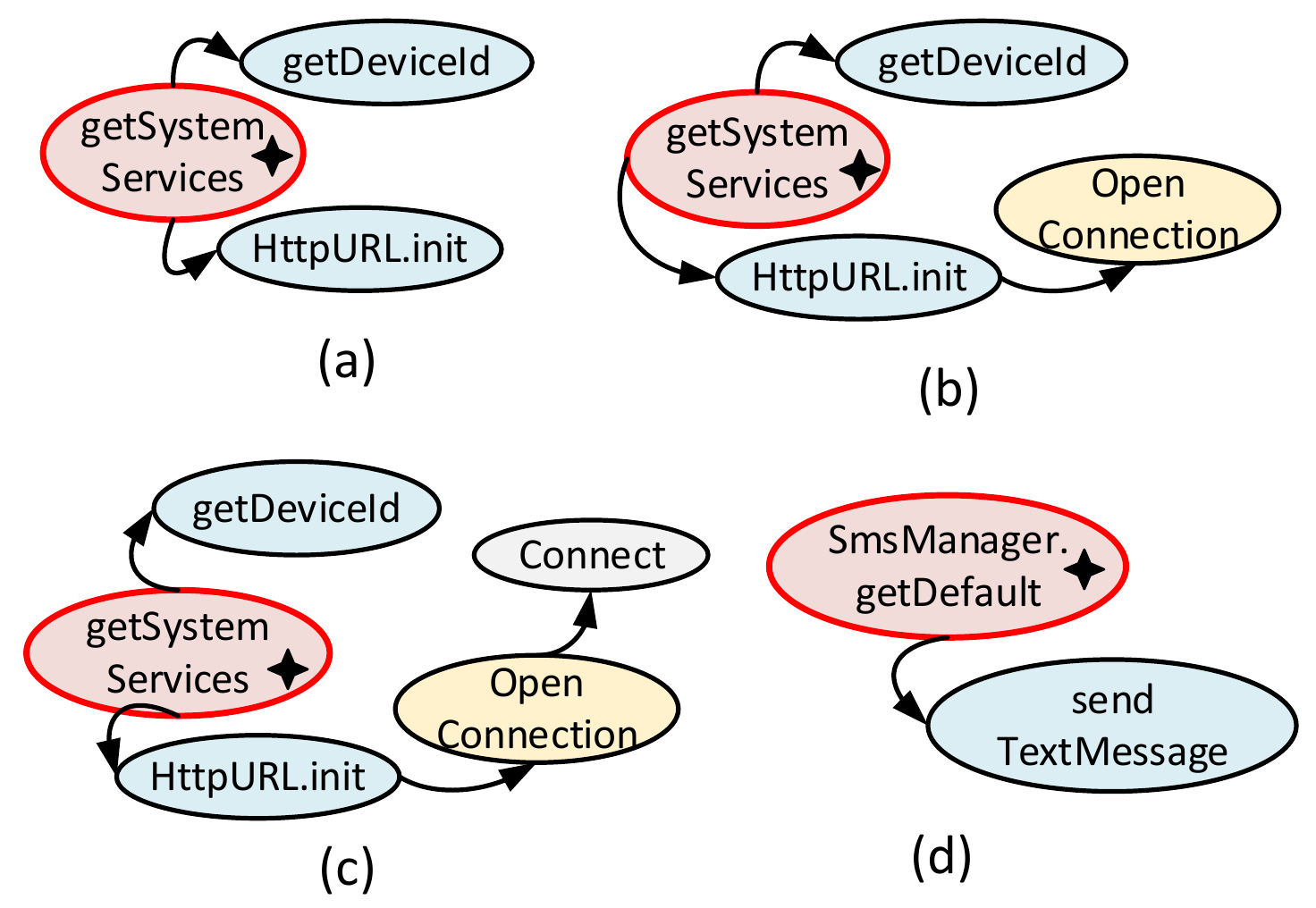}
	\caption{Dependency schema of a set of rooted subgraphs of degree 1 ((a), (d)), 2 ((b)) and 3 ((c)) in an Android malware's API dependency graph. The root nodes are marked with a star. Graph (b) can be derived from (a) by adding a node and an edge. Graph (c) can be derived from (b) in a similar fashion. Graph (d) is highly dissimilar from all the other graphs and is not readily derivable from any of them.  \label {fig:subst_sim}}
\end{figure}

\subsection{Limitations of Existing Graph Kernels}
\label{ss:gk_lim}
However, as noted in \cite{dgk,gksmoothing}, the representation in eq. (1) does not take two important observations into account. 
\begin{itemize}
	[leftmargin=*]
	\setlength\itemsep{0em}
	\item \textbf{(L1) Substructure Similarity.} Substructures that are used to compute the kernel matrix are not independent. To illustrate this, lets consider the Weisfeiler-Lehman (WL) kernel \cite{WLK}  which decomposes graphs into rooted subgraphs\footnote{The WL kernel models the subgraph around a root node as a tree (i.e., without cycles) and hence is referred as WL subtree kernel. However since the tree represents a rooted subgraph, we refer to the rooted subgraph as the substructure being modeled in WL kernel, in this work.}. These subgraphs encompass the neighbourhood of certain degree around the root node. Understandably, these subgraphs exhibit strong relationships among them. 
	That is, a subgraph with second degree neighbours of the root node could be arrived at by adding a few nodes and edges to its first degree counterpart. 
	We explain this with an example presented in Fig. \ref{fig:subst_sim}. The figure illustrates API-dependency subgraphs from a well-known Android malware called DroidKungFu (DKF) \cite{Drebin}. These subgraph portions of DKF involves in leaking users' private information (i.e., IMEI number) over the internet and sending premium-rates SMS without her consent. Sub-figures (a), (b) and (c) represent subgraphs of degree 1, 2 and 3 around the root node {\tt getSystemServices}, respectively. Evidently, these subgraphs exhibit high similarity among one another. For instance, subgraph (c) could be derived from subgraph (b) by adding a node and an edge, which in turn could be derived from subgraph (a) in a similar fashion.  However, the WL kernel, by design ignores these subgraph similarities and considers each of the subgraphs as individual features. Other kernels such as random walk and shortest path kernels also make similar assumptions on their respective substructures' similarities.
	
	\item \textbf{(L2) Diagonal Dominance.} Since graph kernels regard these substructures as separate features, the dimensionality of the feature space often grows exponentially with the number of substructures. Consequently, only a few substructures will be common across graphs. This leads to diagonal dominance, that is, a given graph is similar to itself but not to any other graph in the dataset. This leads to poor classification/clustering accuracy.
\end{itemize}

\subsection{Existing Solution: Deep Graph Kernels}
\label{ss:dgk_lim}
To alleviate these problems Yanardag and Vishwanathan \cite{dgk}, recently proposed an alternative kernel formulation termed as \textit{Deep Graph Kernel} (DGK). Unlike eq. (1), DGK captures the similarities among the substructures with the following formulation:
\begin{equation}
	\mathcal{K} (G, G') =  \Phi (G)^T \mathcal{M}  \Phi (G')
\end{equation}
where $\mathcal{M}$ represents a $ | \mathcal{V}| \times | \mathcal{V} | $ positive semi-definite matrix that encodes the relationship between substructures and $ \mathcal{V} $ represents the vocabulary of substructures obtained from the training data. Therefore, one can design a $\mathcal{M}$  matrix that respects the similarity of the substructure space. 

\textbf{Learning representation of substructures.} In DGK \cite{dgk}, the authors used representation learning (deep learning) techniques inspired by the work of Mikolov et al. \cite{word2vec} to learn vector representations (\textit{aka} embeddings) of substructures. Subsequently, these substructure embeddings were used to compute $\mathcal{M}$ and the same is used in eq (2) to arrive at the deep learning variants of several well-known kernels such as WL, graphlet and shortest path kernels.

\textbf{Context.} In order to facilitate unsupervised representation learning on graph substructures, the authors of \cite{dgk} defined a notion of \textit{context} among these substructures. Substructures that co-occur in the same context tend to have high similarity. For instance, in the case of rooted subgraphs, all the subgraphs that encompass same degree of neighbourhood around the root node are considered as co-occurring in the same context (e.g., all degree-1 subgraphs are considered to be in the same context). Subsequently, embedding learning task's objective is designed to make the embeddings of substructures that occur in the same context similar to one another. Thus defining the correct context is of paramount importance to build high quality embeddings.

\textbf{Deep WL Kernel.} Through their experiments the authors demonstrated that the deep learning variant of WL kernel constructed using the above-said procedure achieved state-of-the-art performances on several datasets. 
However, we observe that, in their approach to learn subgraph embeddings, the authors make three novice assumptions that lead to three critical problems:
\begin{itemize}
	[leftmargin=*]
	\setlength\itemsep{0em}
	\item (\textbf{A1}) \textit{Only rooted subgraphs of same degree are considered as co-occurring in the same context.} That is, if $ D^{(d)}_G = \{sg^{(d)}_1, sg^{(d)}_2, ... \} $ is a multi-set of all degree $ d $ subgraphs in graph G, \cite{dgk} assumes that any two subgraphs $sg^{(d)}_i,sg^{(d)}_j $ $\in$ $ D^{(d)}_G $ co-occur in the same context irrespective of the length (or number) of path(s) connecting them or whether they share the same nodes/edges. For instance, in the case of Android malware subgraphs in Fig. \ref{fig:subst_sim},  
	\cite{dgk} assumes that only subgraphs (a) and (d) are in the same context and are possibly similar as they both are degree-1 subgraphs. However in reality, they share nothing in common and are highly dissimilar. This assumption makes subgraphs that do not co-occur in the same graph neighbourhood to be in the same context and thus similar (problem 1).
		
	\item (\textbf{A2}) \textit{Any two rooted subgraphs of different degrees never co-occur in the same context.} That is, two subgraphs $sg^{(d)}_i \in D^{(d)}_G$ and $sg^{(d')}_j \in D^{(d')}_G $ (where $ d \neq d' $) never co-occur in the same context irrespective of the length (or number) of path(s) connecting them or whether they share the same nodes/edges. For instance, in Fig. \ref{fig:subst_sim}, subgraphs (a), (b) and (c) are considered not co-occurring in the same context as they belong to different degree neighbourhood around the root node. Hence, \cite{dgk} incorrectly biases them to be dissimilar. This assumption makes subgraphs that co-occur in the same neighbourhood not to be in the same context and thus dissimilar (problem 2). 
	
	\item (\textbf{A3}) \textit{Every subgraph ($ sg_r^{(d)} $) in any given graph has exactly same number of subgraphs in its context.} This assumption clearly violates the topological neighbourhood structure in graphs (problem 3).
	
\end{itemize}

Through our thorough analysis and experiments we observe that these assumptions led \cite{dgk} to building relatively low quality subgraph embeddings. Consequently, this reduces the classification and clustering accuracies when \cite{dgk}'s deep WL kernel is deployed.
This motivates us to address these limitations and build better subgraph embeddings, in order to achieve higher accuracy. 

\subsection{Our Approach}
\label{ss:our_appr}
In order to learn accurate subgraph embeddings, we address each of the three problems introduced in the previous subsection. We make two main contributions through our \tool{} framework to solve these problems:
\begin{itemize}
	[leftmargin=*]
	\setlength\itemsep{0em}
	\item We extend the WL relabeling strategy \cite{WLK} (used to relabel the nodes in a graph encompassing its breadth-first neighbourhood) to define a proper context for a given subgraph. For a given subgraph $ sg_r^{(d)} $ in $ G $ with root $ r $, \tool{} considers all the rooted subgraphs (up to a certain degree) of neighbours of $ r $ as the context of $ sg_r^{(d)} $. This solves problems 1 and 2.
	
	\item However this context formation procedure yields radial contexts of different sizes for different subgraphs. This renders the existing representation learning models such as the skipgram model \cite{word2vec} (which captures fixed-length linear contexts) unusable in a straight-forward manner to learn the representations of subgraphs using its context, thus formed. To address this we propose a modification to the skipgram model enabling it to capture varying length radial contexts. This solves problem 3.
\end{itemize}

\textbf{Experiments.} We determine \tool's accuracy and efficiency in both supervised and unsupervised learning tasks with several benchmark and large-scale real-world datasets. Also, we perform comparative analysis against several state-of-the-art graph kernels.
Our experiments reveal that \tool{} achieves significant improvements in classification/clustering accuracy over existing kernels. Specifically, on two real-world program analysis tasks, namely, code clone and malware detection, \tool{} outperforms state-of-the-art kernels by more than 17\% and 4\%, respectively.

\textbf{Contributions.} We make the following contributions:
\begin{itemize}
	[leftmargin=*]
	\setlength\itemsep{0em}
	\item We propose \tool{}, an unsupervised representation learning technique to learn latent representations of rooted subgraphs present in large graphs (\S \ref{sec:learning}).
	
	\item We develop a modified version of the skipgram language model \cite{word2vec} which is capable of modeling varying length radial contexts (rather than fixed-length linear contexts) around target subgraphs (\S \ref{ss:sg2vec_algo}).
	
	\item We discuss how \tool’s representation learning technique would help to build the deep learning variant of WL kernel (\S \ref{ss:dwlk}).
	
	\item Through our large-scale experiments on several benchmark and real-world datasets,  we demonstrate that \tool{} could significantly outperform state-of-the-art graph kernels (incl. \cite{dgk}) on graph classification and clustering tasks (\S \ref{sec:eval}).

\end{itemize}

\section {Related Work}
\label{sec:rw}
The closest work to our paper is Deep Graph Kernels \cite{dgk}. Since we have discussed it elaborately in \S \ref{sec:intro}, we refrain from discussing it here.
Recently, there has been significant interest from the research community on learning representations of nodes and other substructures from graphs. We list the prominent such works in Table \ref{tab:rw} and show how our work compares to them in-principle. Deep Walk \cite{dw} and node2vec \cite{node2vec} intend to learn node embeddings by generating random walks in a single graph. Both these works rely on existence of node labels for at least a small portion of nodes and take a semi-supervised approach to learn node embeddings. Recently proposed Patchy-san \cite{patchysan} learns node and subgraph embeddings using a supervised convolutional neural network (CNN) based approach. In contrast to these three works, \tool{} learns subgraph embeddings (which includes node embeddings) in an unsupervised manner.

In general, from a substructure analysis point of view, research on graph kernel could be grouped into three major categories:  kernels for limited-size subgraphs \cite{pattern_kernel}, kernels based on subtree patterns\cite{WLK} and kernels based on walks \cite{GKVishy} and paths \cite {sp}. \tool{} is complementary to these existing graph kernels where the substructures exhibit reasonable similarities among them.
\begin{table}[t]
	\centering
	\setlength\tabcolsep{0pt}
	\scriptsize
	\caption{Representation Learning from Graphs}
	\label{tab:rw}
	\begin{tabular}{|l|c|c|c|l|}
		\hline
		\textbf{Solution} & \textbf{\begin{tabular}[c]{@{}c@{}}Learning\\ Paradigm\end{tabular}} & \textbf{\begin{tabular}[c]{@{}c@{}}node\\ vector\end{tabular}} & \textbf{\begin{tabular}[c]{@{}c@{}}subgraph\\ vector\end{tabular}} & \textbf{\begin{tabular}[c]{@{}c@{}}Context used for \\ rep. learning\end{tabular}} \\ \hline
		\textbf{Deep Walk} \cite{dw} & Semi-sup & \pie{360} & \pie{0} & \begin{tabular}[c]{@{}l@{}}Fixed-length\\ random walks\end{tabular} \\ \hline
		\textbf{node2vec} \cite{node2vec} & Semi-sup & \pie{360} & \pie{0} & \begin{tabular}[c]{@{}l@{}}Fixed-Length biased\\ random walks\end{tabular} \\ \hline
		\textbf{Patchy-san} \cite{patchysan} & Sup & \pie{360} & \pie{360} &
		\begin{tabular}[c]{@{}l@{}}Receptive field of sequence\\  of  neighbours of nodes\end{tabular}
		\\ \hline
		\textbf{\begin{tabular}[c]{@{}l@{}}Deep Graph\\ Kernels \end{tabular}} \cite{dgk} & Unsup & \pie{360} & \pie{180} & \begin{tabular}[c]{@{}l@{}}Subgraphs occurring\\ at same degree\end{tabular} \\ \hline
		\textbf{\tool} & Unsup & \pie{360} & \pie{360} & \begin{tabular}[c]{@{}l@{}}Subgraphs of different \\ degrees occurring in the \\same local neighbourhoods\end{tabular} \\ \hline
	\end{tabular}
\end{table}
\section {Problem statement}
\label{sec:bg}
We consider the problem of learning distributed representations of rooted subgraphs from a given set of graphs. More formally, let $G = (V, E, \lambda)$, represent a graph, where $V$ is a set of nodes and $E \subseteq (V\times V) $ be a set of edges. 
Graph $ G $ is labeled\footnote{For graphs without node labels, we follow the procedure mentioned in \cite{WLK} and label nodes with their degree.} if there exists a function $ \lambda $ such that $\lambda: V \rightarrow \ell$, which assigns a unique label from alphabet $ \ell $ to every node $ v \in V $. 
Given $ G = (V, E, \lambda) $ and $ sg = (V_{sg}, E_{sg}, \lambda_{sg}) $, $ sg $ is a sub-graph of $ G $ iff there exists an injective mapping $ \mu: V_{sg} \rightarrow V $ such that $ (v_1, v_2) \in E_{sg} $ iff $ (\mu(v_1), \mu(v_2)) \in E $.

\textit{Given a set of graphs $\mathcal{G} = \{G_1, G_2,...,G_n\}$ and a positive integer D, we intend to extract a vocabulary of all (rooted) subgraphs around every node in every graph $G_i \in \mathcal {G} $ encompassing neighbourhoods of degree $0 \le d \le D $, such that $ SG_{vocab} = \{sg_1, sg_2,...\} $. Subsequently, we intend to learn distributed representations with $ \delta $ dimensions for every subgraph $ sg_i \in SG_{vocab} $. The matrix of representations (embeddings) of all subgraphs is denoted as $ \Phi \in \mathbb{R}^{|SG_{vocab}| \times \delta} $}.

Once the subgraph embeddings are learnt, they could be used to cater applications such as graph classification, clustering, node classification, link prediction and community detection. They could be readily used with classifiers such as CNNs and Recurrent Neural Networks. Besides this, these embeddings could be used to make a graph kernel (as in eq(2)) and subsequently used along with kernel classifiers such as SVMs and relational data clustering algorithms. These use cases are elaborated later in \S \ref{ss:uc} after introducing the representation learning methodology.

\section {Background: Language Models}
Our goal is to learn the distributed representations of subgraphs extending the recently proposed representation learning and language modeling techniques for multi-relational data. In this section, we review the related background in language modeling.

\textbf{Traditional language models.} Given a corpus, the traditional language models determine the likelihood of a sequence of words appearing in it. For instance, given a sequence of  words $ \{w_1, w_2, . . . , w_T \} $, n-gram language model targets to maximize the following probability:
\begin{equation}
Pr (w_t | w_1,...,w_{t-1})
\end{equation}

Meaning, they estimate the likelihood of observing the target word $ w_t $ given $ n $ previous words $ (w_1,...,w_{t-1}) $ observed thus far.

\textbf{Neural language models.} 
The recently developed neural language models focus on learning distributed vector representation of words. These models improve traditional n-gram models by using vector embeddings for words. Unlike n-gram models, neural language models exploit the of the notion of context where a \textit{context is defined as a fixed number of words surrounding the target word}. To this end, the objective of these word embedding models is to maximize the following log-likelihood:
\begin{equation}
	\sum_{t=1}^{T} log Pr (w_t | w_{t-c},...,w_{t+c})
\end{equation}
where $ (w_t | w_{t-c},...,w_{t+c}) $ are the context of the target word $ w_t $.
Several methods are proposed to approximate eq. (4). Next, we discuss one such a method that we extend in our \tool{} framework, namely Skipgram models \cite{word2vec}.

\subsection{Skip Gram} 
\label{ss:bg_sg}
The skipgram model maximizes co-occurrence probability among the words that appear within a given context window. 
Give a context window of size $ c $ and the target word $ w_t $, skipgram model attempts to predict the words that appear in the context of the target word, $ (w_{t-c},...,w_{t-c}) $.
More precisely, the objective of the skipgram model is to maximize the following loglikelihood,
\begin{equation}
\sum_{t=1}^{T} log \ Pr (w_{t-c},...,w_{t+c}  | w_t)
\end{equation}
where the probability $ Pr (w_{t-c},...,w_{t+c})$ is computed as
\begin{equation}
\Pi_{-c \le j \le c, j \ne 0} Pr (w_{t+j} | w_t)
\end{equation}

Here, the contextual words and the current word are assumed to be independent. Furthermore, $ Pr (w_{t+j} | w_t) $ is defined as:
\begin{equation}
	\frac {exp(\Phi_{w_t}^T \Phi_{w_{t+j}}^{'})} {\sum_{w=1}^\mathcal{V} exp(\Phi_{w_t}^T \Phi_{w}^{'})}
\end{equation}

 where $ \Phi_{w} $ and $ \Phi_{w}{'} $ are the input and output vectors of word $ w $.

 \subsection{Negative Sampling}
 \label{ss:bg_ns}
 The posterior probability in eq. (6) could be learnt in several ways. For instance, a novice approach is to use a classifier like logistic regression. This is prohibitively expensive if the vocabulary of words is very large.
 
 Negative sampling is an efficient algorithm that is used to alleviate this problem and train the skipgram model. Negative sampling selects the words that are not in the context at random instead of considering all words in the vocabulary. In other words, if a word $ w $ appears in the context of another word $ w' $, then the vector embedding of $ w $ is closer to that of $ w' $ compared to any other randomly chosen word from the vocabulary.
  
 Once skipgram training converges, semantically similar words are mapped to closer positions in the embedding space revealing that the learned word embeddings preserve semantics. 
 An important intuition we extend in \tool{} is to view subgraphs in large graphs as words that are generated from a special language. In other words, different subgraphs compose graphs in a similar way that different words form sentences when used together. With this analogy, one can utilize word embedding models to learn dimensions of similarity between subgraphs. The main expectation here is that similar subgraphs will be close to each other in the embedding space.

\section {Method: Learning sub-graph representations}
\label{sec:learning}
\begin{algorithm}[t]
	\scriptsize
	\caption{ \textsc{subgraph2vec} ($ \mathcal{G}, D, \delta, \mathfrak{e}$) \label{algo:sg2vec}}
	\SetKwInOut{Input}{input}
	\SetKwInOut{Output}{output}
	\Input{$\mathcal{G} = \{G_1, G_2,...,G_n\}$: set of graphs such that each graph $ G_i = (V_i, E_i, \lambda_i) $ from which embeddings are learnt\newline
		$ D $: Maximum degree of subgraphs to be considered for learning representations. This will produce a vocabulary of subgraphs, $ SG_{vocab} = \{sg_1, sg_2, ...\} $ from all the graphs in $ \mathcal{G} $\newline
		$\delta$: number of dimensions (embedding size)\newline
		$ \mathfrak{e} $: number of epochs}
	\Output{Matrix of vector representations of subgraphs $ \Phi \in \mathbb{R}^{|SG_{vocab}| \times \delta}$}

	\small
	\Begin{
		$ SG_{vocab} $ = \textsc{BuildSubgraphVocab}($ \mathcal{G} $) \textcolor{gray}{//use Algorithm 2}\\
		Initialization: Sample $ \Phi $ from $ U^{|SGVocab| \times \delta} $\\		
		\For{$ e = 0$ to $\mathfrak{e} $}{
			$ \mathfrak{G} $ = \textsc{Shuffle} ($ \mathcal{G} $)\\
			\For {\textbf{each} $G_i \in \mathfrak{G}$}{
				\For {\textbf{each} $ v \in V_i $}{
					\For {$ d = 0 $ to $ D $}{
						$ sg^{(d)}_{v} := $ \textsc{GetWLSubgraph}($v, G_i, d $)\\
							\textsc{RadialSkipGram} ($ \Phi, sg^{(d)}_{v}, G_i, D$)
						}
					}
				}
		}
		\textbf{return} $\Phi$
	}	
\end{algorithm}

In this section we discuss the main components of our \tool{} algorithm (\S \ref{ss:sg2vec_algo}), how it enables making a deep learning variant of WL kernel (\S \ref{ss:dwlk}) and some of its usecases in detail (\S \ref{ss:uc}).

\subsection {Overview}
\label{ss:ov}
Similar to the language modeling convention, the only required input is a corpus and a vocabulary of subgraphs for \tool{} to learn representations.  Given a dataset of graphs, \tool{} considers all the neighbourhoods of rooted subgraphs around every rooted subgraph (up to a certain degree) as its \textit{corpus}, and set of all rooted subgraphs around every node in every graph as its \textit{vocabulary}. Subsequently, following the language model training process with the subgraphs and their contexts, \tool{} learns the intended subgraph embeddings. 

\subsection{Algorithm: \tool} 
\label{ss:sg2vec_algo}
The algorithm consists of two main components; first a procedure to generate rooted subgraphs around every node in a given graph (\S \ref {sss:subgraph_ext}) and second the procedure to learn embeddings of those subgraphs (\S \ref{sss:radial_sg}). 

As presented in Algorithm \ref{algo:sg2vec} we intend to learn $ \delta $ dimensional embeddings of subgraphs (up to degree $ D $) from all the graphs in dataset $ \mathcal{G} $ in $ \mathfrak{e} $ epochs. 
We begin by building a vocabulary of all the subgraphs, $ SG_{vocab} $  (line 2) (using Algorithm \ref{algo:get_rooted_sg}). Then the embeddings for all subgraphs in the vocabulary ($ \Phi $) is initialized randomly (line 3). Subsequently, we proceed with learning the embeddings in several epochs (lines 4 to 10) iterating over the graphs in $ \mathcal{G} $. These steps represent the core of our approach and are explained in detail in the two following subsections.

\begin{algorithm}[t]
	\scriptsize
	\caption{\textsc{GetWLSubgraph} $ (v, G, d) $}
	\label{algo:get_rooted_sg}
	\SetKwInOut{Input}{input}
	\SetKwInOut{Output}{output}
	\Input{$v $: Node which is the root of the subgraph  \newline
		$ G = (V,E,\lambda) $: Graph from which subgraph has to be extracted \newline
		$d$: Degree of neighbours to be considered for extracting subgraph }
	\Output{$sg^{(d)}_v$: rooted subgraph of degree $ d $ around node $ v $ }
	
	\small
	\Begin{
		$ sg^{(d)}_v $ = \{\} \newline
		\If {$d = 0$} {
			$sg^{(d)}_v := \lambda(v) $ 
		}
		\Else{
			$\mathcal{N}_v := \{v'\ |\ (v,v') \in E\}$\\
			$M^{(d)}_v := \{$\textsc{GetWLSubgraph}$(v',G,d-1)\ |\ v' \in \mathcal{N}_v \}$ \\
			$sg^{(d)}_v := sg^{(d)}_v \cup $  \textsc{GetWLSubgraph} $(v,G,d-1) \oplus sort(M^{(d)}_v)$  
		}
		\textbf{return } $ sg^{(d)}_v $
	}
\end{algorithm}

\begin{algorithm}[t]
	\small
	\caption{\textsc{RadialSkipGram} ($\Phi, sg_{v}^{(d)}, G, D$) }
	\label{algo:skipgram}	
	\Begin{
		$ context^{(d)}_{v} = \{\} $\\
		\For {$ v' \in $ \textsc{Neighbours}($ G, v $)}{
			\For {$ \partial \in \{d-1,\ d,\ d+1\} $}{
				\If  {($ \partial \ge 0 $ \textbf{and} $ \partial \le D $)}{
					$ context^{(d)}_{v} $ = $ context^{(d)}_{v} \cup $ \textsc{GetWLSubgraph}($v', G, \partial$)}}
		}
		\For {\textbf{each} $ sg_{cont} \in context^{(d)}_{v} $}
		{
			$ J (\Phi) $ = $ - $log Pr ($ sg_{cont} | \Phi(sg_{v}^{(d)}) $)\\
			$ \Phi = \Phi - \alpha  \frac{\partial J}{\partial \Phi} $
		}
	}
\end{algorithm}
 
\subsubsection{Extracting Rooted Subgraphs}
\label{sss:subgraph_ext}
To facilitate learning its embeddings, a rooted subgraph $ sg_{v}^{(d)} $ around every node $ v $ of graph $ G_i $ is extracted (line 9). This is a fundamentally important task in our approach. To extract these subgraphs, we follow the well-known WL relabeling process \cite{WLK} which lays the basis for the WL kernel and WL test of graph isomorphism \cite{dgk,WLK}. The subgraph extraction process is explained separately in Algorithm \ref{algo:get_rooted_sg}. The algorithm takes the root node $ v $, graph $ G $ from which the subgraph has to be extracted and degree of the intended subgraph $ d $ as inputs and returns the intended subgraph $ sg_v^{(d)} $. When $ d =0 $, no subgraph needs to be extracted and hence the label of node $ v $ is returned (line 3). For cases where $ d > 0 $, we get all the (breadth-first) neighbours of $ v $ in $ \mathcal{N}_v $ (line 5). Then for each neighbouring node, $ v' $, we get its degree $ d-1 $ subgraph and save the same in list $ M^{(d)}_v $ (line 6). Finally, we get the degree $ d-1 $ subgraph around the root node $ v $ and concatenate the same with sorted list $ M_v^{(d)} $ to obtain the intended subgraph $ sg_v^{(d)} $ (line 7).  

\textbf{Example.} To illustrate the subgraph extraction process, lets consider the examples in Fig. \ref{fig:subst_sim}. Lets consider the graph 1(c) as the complete graph from which we intend to get the degree 0, 1 and 2 subgraph around the root node {\tt HttpURL.init}. Subjecting these inputs to Algorithm \ref{algo:get_rooted_sg}, we get subgraphs \{{\tt HttpURL.init}\}, \{{\tt HttpURL.init -> OpenConnection}\} and \{{\tt HttpURL.init -> OpenConnection -> Connect}\} for degrees 0, 1 and 2, respectively.

\subsubsection {Radial Skipgram}
\label{sss:radial_sg}

Once the subgraph $ sg_v^{(d)} $, around the root node $ v $ is extracted, Algorithm \ref{algo:sg2vec} proceeds to learn its embeddings with the radial skip gram model (line 10).
Similar to the vanilla skipgram algorithm which learns the embeddings of a target word from its surrounding linear context in a given document, our approach learns the embeddings of a target subgraph using its surrounding radial context in a given graph. The radial skipgram procedure is presented in Algorithm \ref{algo:skipgram}. 

\textbf{Modeling the radial context.}
 The radial context around a target subgraph is obtained using the process explained below. As discussed previously in \S \ref{ss:bg_sg}, natural language text have linear co-occurrence relationships. For instance, skipgram model iterates over all possible collocations of words in a given sentence and in each iteration it considers one word in the sentence as the \textit{target word} and the words occurring in its context window as \textit{context words}.
This is directly usable on graphs if we model linear substructures such as walks or paths with the view of building node representations. For instance, Deep Walk \cite{dw} uses a similar approach to learn a target node's representation by generating random walks around it. However, unlike words in a traditional text corpora, subgraphs do not have a linear co-occurrence relationship. Therefore, we intend to consider the breadth-first neighbours of the root node as its context as it directly follows from the definition of WL relabeling process. 

To this end, we define the context of a degree-$ d $ subgraph $ sg_{v}^{(d)} $ rooted at $ v $, as the multiset of subgraphs of degrees $ d-1, d $ and $ d+1 $ rooted at each of the neighbours of $ v $ (lines 2-6 in Algorithm \ref{algo:skipgram}). Clearly this models a radial context rather than a linear one. Note that we consider subgraphs of degrees $ d-1, d $ and $ d+1 $ to be in the context of a subgraph of degree $ d $. This is because, as explained with example earlier in \S \ref{ss:gk_lim}, a degree-$ d $ subgraph is likely to be rather similar to subgraphs of degrees that are closer to $ d $ (e.g., $ d-1, d+1 $) and not just degree-$ d $ subgraphs only.
  
\textbf{Vanilla Skip Gram.} 
As explained previously in \S \ref{ss:bg_sg}, the vanilla skipgram language model captures fixed-length linear contexts over the words in a given sentence.
However, for learning a subgraph's radial context arrived at line 6 in Algorithm \ref{algo:skipgram}, the vanilla skipgram model could not be used. Hence we propose a minor modification to consider a radial context as explained below. 

\textbf{Modification.}
The embedding of a target subgraph, $sg^{(d)}_{v}$, with context $ context^{(d)}_{v} $ is learnt using lines 7 - 9 in Algorithm \ref{algo:skipgram}. 
Given the current representation of target subgraph $\Phi (sg^{(d)}_{v})$, we would like to maximize the probability of every subgraph in its context $ sg_{cont} $ (lines 8 and 9). We can learn such posterior distribution using several choices of classifiers.
For example, modeling it using logistic regression would result in a huge number of labels that is equal to $ | SG_{vocab} | $. This could be in several thousands/millions in the case of large graphs. Training such models would require large amount of computational resources. 
To alleviate this bottleneck, we approximate the probability distribution using the negative sampling approach.


\subsubsection{Negative Sampling}
\label{sss:ns}
Given that $sg_{cont} \in SG_{vocab}$ and $ |SG_{vocab}| $ is very large, calculating $ Pr (sg_{cont} | \Phi(sg^{(d)}_{v}))$ in line 8 is prohibitively expensive. Hence we follow the negative sampling strategy (introduced in \S \ref{ss:bg_ns}) to calculate above mentioned posterior probability.
In our negative sampling phase for every training cycle of Algorithm \ref{algo:skipgram}, we choose a fixed number of subgraphs (denoted as $ negsamples $) as negative samples and update their embeddings as well. Negative samples adhere to the following conditions: if $negsamples = \{sgneg_1, sgneg_2, ...\} $, then { \small $ negsamples \subset SG_{vocab}$, $|negsamples| << |SG_{vocab}|$ and $negsamples \ \cap\ context^{(d)}_{v} = \{\}$}.
This makes $\Phi(sg^{(d)}_{v})$ closer to the embeddings of all the subgraphs its context (i.e.$\Phi (sg_{cont})\ | \ \forall sg_{cont} \in context^{(d)}_{v}$) and at the same time distances the same from  the embeddings of a fixed number of subgraphs that are not its context (i.e.$\Phi (sgneg_i)\ |\  \forall sgneg_i \in negsamples$).
\subsubsection{Optimization}
\label{sss:opt}
Stochastic gradient descent (SGD) optimizer is used to optimize these parameters (line 9, Algorithm \ref{algo:skipgram}). The derivatives are estimated using the back-propagation algorithm. The learning rate $ \alpha $ is empirically tuned.

\subsection {Relation to Deep WL kernel}
\label{ss:dwlk}
As mentioned before, each of the subgraph in $SG_{vocab}$ is obtained using the WL re-labelling strategy, and hence represents the WL neighbourhood labels of a node. Hence learning latent representations of such subgraphs amounts to learning representations of WL neighbourhood labels. Therefore, once the embeddings of all the subgraph in $SG_{vocab}$ are learnt using Algorithm \ref{algo:sg2vec}, one could use it to build the deep learning variant of the WL kernel among the graphs in $\mathcal{G}$. 
For instance, we could compute $ \mathcal{M} $ matrix such that each entry $ \mathcal{M}_{ij} $ computed as $ \langle \Phi_i, \Phi_j \rangle $ where $ \Phi_i $ corresponds to learned $ \delta $-dimensional embedding of  subgraph $ i $ (resp. $ \Phi_j $ ).
Thus, matrix $\mathcal{M}$ represents nothing but the pairwise similarities of all the substructures used by the WL kernel. Hence, matrix $\mathcal{M}$ could directly be plugged into eq. (2)  to arrive at the deep WL kernel across all the graphs in $\mathcal{G}$.

\subsection{Use cases}
\label{ss:uc}
Once we compute the subgraph embeddings, they could be used in several practical applications. We list some prominent use cases here: 

\textit{(1) Graph Classification.} Given $ \mathcal{G} $, a set of graphs and $ Y $, the set of corresponding class labels, graph classification is the task where we learn a model $ \mathcal{H} $ such that $ \mathcal{H}: \mathcal{G} \rightarrow Y $. To this end, one could feed \tool's embeddings to a deep learning classifier such as CNN (as in \cite{patchysan}) to learn $ \mathcal{H} $. Alternatively, one could follow a kernel based classification. That is, one could arrive at a deep WL kernel using the subgraph embeddings as discussed in \S \ref{ss:dwlk}, and use kernelized learning algorithm such as SVM to perform classification.

\textit{(2) Graph Clustering.} Given $ \mathcal{G} $, in graph clustering, the task is to group similar graphs together. Here, a graph kernel 
could be used to calculate the pairwise similarity among graphs in $ \mathcal{G} $. Subsequently, relational data clustering algorithms such as Affinity Propagation (AP) \cite{AP} and Hierarchical Clustering could be used to cluster the graphs.  

It is noted that \tool's use cases are not confined only to the aforementioned tasks. Since \tool{} could be used to learn node representations (i.e., when subgraph of degree 0 are considered, \tool{} provides node embeddings similar to Deep Walk \cite{dw} and node2vec \cite{node2vec}). Hence other tasks such as node classification, community detection and link prediction could also performed using \tool's embeddings. However, in our evaluations in this work we consider only graph classification and clustering as they are more prominent.

\section {Evaluation}
\label{sec:eval}
We evaluate \tool's accuracy and efficiency both in supervised and unsupervised learning tasks. Besides experimenting with benchmark datasets, we also evaluate \tool{} on with real-world program analysis tasks such as malware and code clone detection on large-scale Android malware and clone datasets. 
Specifically, we intend to address the following research questions: (1) How does \tool{} compare to existing graph kernels for graph classification tasks in terms of accuracy and efficiency on benchmark datasets, (2) How does \tool{} compare to state-of-the-art graph kernels on a real-world unsupervised learning task, namely, code clone detection (3) How does \tool{} compare to state-of-the-art graph kernels on a real-world supervised learning task, namely, malware detection.

\textbf{Evaluation Setup.}
All the experiments were conducted on a server with 36 CPU cores (Intel E5-2699 2.30GHz processor), NVIDIA GeForce GTX TITAN Black GPU and 200 GB RAM running Ubuntu 14.04.
\begin{table}[t]
	\centering
	\setlength\tabcolsep{4pt}
	\scriptsize
	\caption{Benchmark dataset statistics}
	\label{tab:bm_ds}
	\begin{tabular}{|c|c|c|c|}
		\hline
		\textbf{Dataset}  & \textbf{\# samples} & \textbf{\begin{tabular}[c]{@{}l@{}}\# nodes\\ (avg.)\end{tabular}} & \textbf{\begin{tabular}[c]{@{}l@{}}\# distinct \\ node labels\end{tabular}} \\ \hline
		\textbf{MUTAG}    & 188                 & 17.9                     & 7  \\ 
		\textbf{PTC}      & 344                 & 25.5                     & 19\\ 
		\textbf{PROTEINS} & 1113                & 39.1                     & 3 \\ 
		\textbf{NCI1}     & 4110                & 29.8                     & 37 \\ 
		\textbf{NCI109}   & 4127                & 29.6                     & 38  \\ \hline
	\end{tabular}
\end{table}

\begin{table*}[t]
	\setlength\tabcolsep{8 pt}
	\centering
	\scriptsize
	\caption{Average Accuracy ($\pm$ std dev.) for \tool{} and state-of-the-art graph kernels on benchmark graph classification datasets}
	\label{tab:res_bm}
	\begin{tabular}{|l|l|l|l|l|l|}
		\hline
		\textbf{Dataset}                     & \textbf{MUTAG} & \textbf{PTC} & \textbf{PROTEINS} & \textbf{NCI1} & \textbf{NCI109} \\ \hline
		\textbf{WL \cite{WLK}}                          & 80.63 $\pm$ 3.07   & 56.91 $\pm$ 2.79 & 72.92 $\pm$ 0.56     & 80.01 $\pm$ 0.50  & 80.12 $\pm$ 0.34   \\ 
		\textbf{Deep WL\textsubscript{YV} \cite{dgk}}                     & 82.95  $\pm$ 1.96    & 59.04 $\pm$ 1.09  & 73.30 $\pm$ 0.82    & \textbf{80.31} $\pm$ 0.46  & \textbf{80.32} $\pm$ 0.33  \\ 
		\textbf{\tool{}}                & \textbf{87.17} $\pm$ 1.72  & \textbf{60.11} $\pm$ 1.21  & \textbf{73.38} $\pm$ 1.09  & 78.05 $\pm$ 1.15 & 78.39 $\pm$ 1.89 \\ \hline
	\end{tabular}
\end{table*}
\subsection {Classification on benchmark datasets}
\label{ss:eval_bm}
\textbf{Datasets.} Five benchmark graph classification datasets namely MUTAG, PTC, PROTEINS, NCI1 and NCI109 are used in this experiment. These datasets belong to chemo- and bio-informatics domains and the statistics on the same are reported in Table \ref{tab:bm_ds}. 
MUTAG dataset consists 188 chemical compounds where class label indicates whether or not the compound has a mutagenic effect on a bacterium. 
PTC dataset comprises of 344 compounds and the classes indicate carcinogenicity on female/male rats.
PROTEINS is a graph collection where nodes are secondary
structure elements and edges indicate neighborhood in the
amino-acid sequence or in 3D space.
NCI1 and NCI109 datasets contain compounds screened for activity against non-small cell lung cancer and ovarian cancer cell lines.  Graphs are classified as enzyme or non-enzyme. 
All these datasets are made available in \cite{WLK,dgk}. 

\textbf{Comparative Analysis.}
For classification tasks on each of the datasets, we use the embeddings learnt using \tool{} and build the Deep WL kernel as explained in \S \ref{ss:dwlk}.  
We compare \tool{} against the WL kernel \cite{WLK} and Yanardag and Vishwanathan's formulation of deep WL kernel \cite{dgk} (denoted as Deep WL\textsubscript{YV}). 

\textbf{Configurations.} For all the datasets, 90\% of samples are chosen at random for training and the remaining 10\% samples are used for testing. The hyper-parameters of the classifiers are tuned based on 5-fold cross validation on the training set.

\textbf{Evaluation Metric.} The experiment is repeated 5 times and the average accuracy (along with std. dev.) is used to determine the effectiveness of classification. Efficiency is determined in terms of time consumed for learning subgraph embeddings (\textit{aka} pre-training duration).

\subsubsection{Results and Discussion.}
\textbf{Accuracy. }Table \ref{tab:res_bm} lists the results of the experiments.  It is clear that SVMs with \tool’s embeddings achieve better accuracy on 3 datasets (MUTAG,  PTC and PROTEINS) and comparable accuracy on the remaining  2 datasets (NCI1 and NCI109). 

\textbf{Efficiency.} Out of the methods compared, only Deep WL\textsubscript{YV} kernel and \tool{} involve pre-training to compute vectors of subgraphs. Evidently, pre-training helps them capture latent similarities between the substructures in graphs and thus aids them to outperform traditional graph kernels. Therefore, it is important to study the cost of pre-training. To this end, we report the pre-training durations of these two methods in Fig. \ref{fig:time}. Being similar in terms of pre-training, both methods require very similar durations to build the pre-trained vectors. However, for the datasets under consideration, \tool{} requires lesser time than Deep WL\textsubscript{YV} kernel as its radial skipgram involves slightly lesser computations
than the vanilla skipgram used in Deep WL\textsubscript{YV} kernel.

However it is important to note that classification on these benchmark datasets are much simpler than real-world classification tasks. In fact, by using trivial features such as number of nodes in the graph, \cite{gk_ds} achieved comparable accuracies to the state-of-the-art graph kernels. It would be incomplete if we evaluate \tool{} only on these benchmark datasets. Hence in the two subsequent experiments, we involve real-world datasets on practical graph clustering and classification tasks.

\begin{figure}[t]
	\centering
	\includegraphics[height=4cm,width=9cm]{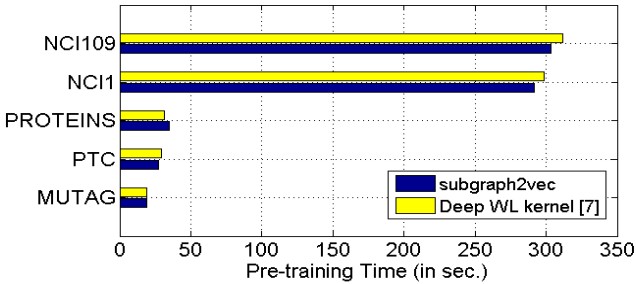}
	\caption{Deep WL kernel Vs subgraph2vec Pre-training Durations. \label {fig:time}}
\end{figure}

\begin{table}[t]
	\centering
	\setlength\tabcolsep{3pt}
	\scriptsize
	\caption{Clone Detection Dataset Statistics}
	\label{tab:clone_ds}
	\begin{tabular}{|c|c|c|c|c|}
		\hline
		\textbf{Dataset}  & \textbf{\# samples} & \textbf{\# clusters} & \textbf{\begin{tabular}[c]{@{}c@{}}\# nodes\\ (avg.)\end{tabular}} & \textbf{{\begin{tabular}[c]{@{}c@{}}\# edges \\ (avg.) \end{tabular}}} \\ \hline
		\textbf{$\textsc{Clone}_{260}$ \cite{3dcfg}}    & 260 & 100 & 9829.15   & 31026.30 \\ \hline
	\end{tabular}
\end{table}
\subsection{Clone Detection}
\label{ss:eval_clone}
Android apps are cloned across different markets by unscrupulous developers for reasons such as stealing advertisement revenue \cite{3dcfg}. Detecting and removing such cloned apps is an important task for app market curators that helps maintaining quality of markets and app ecosystem.
In this experiment, we consider a set of Android apps and our goal is to cluster them such that clone (semantically similar) apps are grouped together. Hence, this amounts to unsupervised code similarity detection.

\begin{table}[t]
	\setlength\tabcolsep{1pt}
	\centering
	\scriptsize
	\caption{Clone Detection - Results}
	\label{tab:clone_res}
	\begin{tabular}{|c|c|c|c|}
		\hline
		\textbf{Kernel} & \textbf{WL \cite{WLK} } & \textbf{Deep WL\textsubscript{YV} \cite{dgk} } & \textbf{subgraph2vec} \\ \hline
		\textbf{\begin{tabular}[c]{@{}l@{}}Pre-training duration\end{tabular}} & - & 421.7 s & 409.28 s\\ 
		\textbf{ARI} & 0.67 & 0.71 & \textbf{0.88} \\ \hline
	\end{tabular}
\end{table}

\textbf{Dataset.} We acquired a dataset of 260 apps collected from the authors of a recent clone detection work, 3D-CFG \cite{3dcfg}. We refer to this dataset as {$\textsc{Clone}_{260}$}. All the apps  in $\textsc{Clone}_{260}$ are manually analyzed and 100 clone sets (i.e. ground truth clusters) are identified by the authors of \cite{3dcfg}. The details on this dataset are furnished in Table \ref{tab:clone_ds}. As it could be seen from the table, this problem involves graphs that are much larger/denser than the benchmark datasets used in \S \ref{ss:eval_bm}.

Our objective is to reverse engineer these apps, obtain their bytecode and represent the same as graphs. Subsequently, we cluster similar graphs that represent cloned apps together.
To achieve this, we begin by representing reverse engineered apps as Inter-procedural Control Flow Graphs (ICFGs). Nodes of the ICFGs are labeled with Android APIs that they access\footnote{For more details on app representations, we refer to \cite{appcontext}.}. Subsequently, we use \tool{} to learn the vector representations of subgraphs from these ICFGs and build a deep kernel matrix (using eq. (2)). Finally, we use AP clustering algorithm \cite{AP} over the kernel matrix to obtain clusters of similar ICFGs representing clone apps. 

\textbf{Comparative Analysis.} We compare \tool's accuracy on the clone detection task against the WL \cite{WLK} and Deep WL\textsubscript{YV} \cite{dgk} kernels. 

\textbf{Evaluation Metric.} A standard clustering evaluation metric, namely, Adjusted Rand Index (ARI) is used to determine clone detection accuracy. The ARI values lies in the range [-1, 1]. A higher ARI means a higher correspondence to ground-truth clone sets.

\subsubsection{Results and Discussion.}
\textbf{Accuracy. }The results of clone detection using the three kernels under discussion are presented in Table \ref{tab:clone_res}. 
Following observations are drawn from the table:
\begin{itemize}
	[leftmargin=*]
	\setlength\itemsep{0em}
	\item \tool{} outperform WL and Deep WL\textsubscript{YV} kernels by more than 21\% and 17\% , respectively. The difference between using Deep WL kernel and \tool{} embeddings is more pronounced in the unsupervised learning task.
	
	\item WL kernel perform poorly in clone detection task as it, by design, fails to identify the subgraph similarities, which is essential to precisely captures the latent program semantics. On the other hand, Deep WL\textsubscript{YV} kernel performs reasonable well as it captures similarities among subgraphs of same degree. However, it fails to capture the complete semantics of the program due to its strong assumptions (see \S \ref{ss:dgk_lim}). Whereas, \tool{} was able to precisely capture subgraph similarities spanning across multiple degrees.
\end{itemize}

\textbf{Efficiency.} From Table \ref{tab:clone_res}, it can be seen that the pretraining duration for subgraph2vec is slightly better than Deep WL\textsubscript{YV} kernel. This observation is inline with the pretraining durations of benchmark datasets. 
WL kernel involves no pre-training and deep kernel computation and hence much more efficient than the other two methods. 

\subsection{Malware Detection}
\label{ss:eval_mal}
Malware detection is a challenging task in the field of cyber-security as the attackers continuously enhance the sophistication of malware to evade novel detection techniques. In the case of Android platform, many existing works such as \cite{appcontext}, represent benign and malware apps as ICFGS and cast malware detection as a graph classification problem. Similar to clone detection, this task typically involves large graphs as well.  \\
\textbf{Datasets.}
\begin{table}[t]
	\setlength\tabcolsep{1pt}
	\centering
	\scriptsize
	\caption{Malware Detection Dataset Statistics}
	\label{tab:mal_ds}
	\begin{tabular}{|c|c|c|c|c|c|}
		\hline
		\textbf{Dataset}        & \textbf{Class} & \textbf{Source}        & \textbf{\# apps} & \textbf{{\begin{tabular}[c]{@{}c@{}}\# nodes \\ (avg.) \end{tabular}}}  & \textbf{{\begin{tabular}[c]{@{}c@{}}\# edges \\ (avg.) \end{tabular}}}  \\ \hline
		\multirow{2}{*}{\textbf{$\textsc{Train}_{10K}$}} & Malware        & Drebin \cite{Drebin} & 5600  & 9590.23 &  19377.96             \\ 
		& Benign         & Google Play \cite{GP}           & 5000   & 20873.71 &  38081.24             \\ \hline
		\multirow{2}{*}{\textbf{$\textsc{Test}_{10K}$}} & Malware        & Virus Share  \cite{VS}          & 5000      & 13082.40 &  25661.93         \\ 
		& Benign         & Google Play  \cite{GP}          & 5000      & 27032.03 &  42855.41         \\ \hline
	\end{tabular}
\end{table}
\textsc{Drebin} \cite{Drebin} provides a collection of 5,560 Android malware apps collected from 2010 to 2012. We collected 5000 benign top-selling apps from Google Play \cite{GP} that were released around the same time and use them along with the Drebin apps to train the malware detection model. We refer to this dataset as {\tt $\textsc{Train}_{10K}$}. To evaluate the performance of the model, we use a more recent set of 5000 malware samples (i.e., collected from 2010 to 2014) provided by Virus share \cite{VS} and an equal number of benign apps from Google Play that were released around the same time. We refer to this dataset as {\tt $\textsc{Test}_{10K}$}. Hence, in total, our malware detection experiments involve 20,600 apps. The statistics of this dataset is presented in Table \ref{tab:mal_ds}.\\
\textbf{Comparative Analysis and Evaluation Metrics.} The same type of comparative analysis and evaluation metrics against WL and Deep WL\textsubscript{YV} kernels used in experiments with benchmark datasets in \S \ref{ss:eval_bm} are used here as well.
\subsubsection{Results \& Discussion.}

\textbf{Accuracy.} The results of malware detection using the
three kernels under discussion are presented in Table \ref{tab:mal_res}.
Following observations are drawn from the table:
\begin{itemize}
	[leftmargin=*]
	\setlength\itemsep{0em}

	\item SVM built using subgraph2vec embeddings outperform WL
	and Deep WL\textsubscript{YV} kernels by more than 12\% and 4\%, respectively. This improvement could be attributed to \tool{}'s high quality embeddings learnt from apps' ICFGs. 
	
	\item On this classification task, both Deep WL\textsubscript{YV} and \tool{} outperform WL kernel by a significant margin (unlike the experiments on benchmark datasets). Clearly, this is due to the fact that the former methods capture the latent subgraph similarities from ICFGs which helps them learn semantically similar but syntactically different malware features.
		
\end{itemize}

\textbf{Efficiency.} 
The inferences on pre-training efficiency discussed in \S \ref{ss:eval_bm} and \S \ref{ss:eval_clone} hold for this experiment as well. 

\label{sss:eval_mal_rd}
\begin{table}[t]
	\setlength\tabcolsep{1pt}
	\centering
	\scriptsize
	\caption{Malware Detection - Results}
	\label{tab:mal_res}
	\begin{tabular}{|c|c|c|c|}
		\hline
		\textbf{Classifier} & \textbf{WL \cite{WLK} } & \textbf{Deep WL\textsubscript{YV} \cite{dgk} } & \textbf{subgraph2vec} \\ \hline
		\textbf{\begin{tabular}[c]{@{}l@{}}Pre-training duration\\ \end{tabular}} & - & 2631.17 s & 2219.28 s\\ 
		\textbf{Accuracy} & 66.15 & 71.03 & \textbf{74.48} \\ \hline
	\end{tabular}
\end{table}

\section{Conclusion}
In this paper, we presented \tool, an unsupervised representation learning technique to learn embedding of rooted subgraphs  that exist in large graphs. 
Through our large-scale experiments involving benchmark and real-world graph classification and clustering datasets, we demonstrate that subgraph embeddings learnt by our approach could be used in conjunction with classifiers such as CNNs, SVMs and relational data clustering algorithms to achieve significantly superior accuracies. On real-world application involving large graphs, \tool{} outperforms state-of-the-art graph kernels significantly without compromising efficiency of the overall performance. We make all the code and data used within this work available at: {\scriptsize \url{https://sites.google.com/site/subgraph2vec}}


\end{document}